\documentclass[10pt,twocolumn,letterpaper]{article}

\usepackage{cvpr}
\usepackage{times}
\usepackage{epsfig}
\usepackage{graphicx}
\usepackage{amsmath}
\usepackage{amssymb}
\usepackage{subcaption}
\usepackage{caption}
\usepackage{bbm}
\usepackage{lipsum}

\usepackage[breaklinks=true,bookmarks=false]{hyperref}

\cvprfinalcopy 

\def\cvprPaperID{****} 

\begin{document}

\title{Boundary-sensitive Network for Portrait Segmentation}

\author{Xianzhi Du$^1$, Xiaolong Wang$^2$, Dawei Li$^2$, Jingwen Zhu$^2$, Serafettin Tasci$^2$, \\
Cameron Upright$^2$, 	Stephen Walsh$^2$, Larry Davis$^1$\\
$^1$Computer Vision Lab, UMIACS, University of Maryland, College Park, 20742\\
$^2$Samsung Research America, Mountain View, 94043\\
{\tt\small \{xianzhi,lsd\}@umiacs.umd.edu}\\
{\tt\small \{xiaolong.w,dawei.l,jingwen.z,s.tasci,c.upright,s1.walsh\}@samsung.com}
}

\maketitle

\begin{abstract} 
Compared to the general semantic segmentation problem, portrait segmentation has higher precision requirement on boundary area. However, this problem has not been well studied in previous works. In this paper, we propose a boundary-sensitive deep neural network (BSN) for portrait segmentation. BSN introduces three novel techniques. First, an individual boundary-sensitive kernel is proposed by dilating the contour line and assigning the boundary pixels with multi-class labels. Second, a global boundary-sensitive kernel is employed as a position sensitive prior to further constrain the overall shape of the segmentation map. Third, we train a boundary-sensitive attribute classifier jointly with the segmentation network to reinforce the network with semantic boundary shape information. We have evaluated BSN on the current largest public portrait segmentation dataset,~\ie, the PFCN dataset, as well as the portrait images collected from other three popular image segmentation datasets: COCO, COCO-Stuff, and PASCAL VOC. Our method achieves the superior quantitative and qualitative performance over state-of-the-arts on all the datasets, especially on the boundary area.
\end{abstract}

\section{Introduction}
Semantic segmentation is a fundamental problem in computer vision community which aims to classify pixels into semantic categories. In this paper, we target a special binary class semantic segmentation problem, namely portrait segmentation, which generates pixel-wise predictions as foreground (\ie, people) or background. Recently, it is becoming a hot topic and has been widely used in many real-world applications, such as augmented reality (AR), background replacement, portrait stylization, depth of field, advanced driver assistance systems~\cite{xz_auto}, etc. Although numerous deep learning based approaches (\eg,~\cite{ChenPSA17} ~\cite{BadrinarayananK15}~\cite{PengZYLS17}~\cite{ZhaoSQWJ16}~\cite{LinMS016}~\cite{dw_1}~\cite{dw_2}) were proposed to solve the general semantic segmentation problem, direct adaptation of these methods cannot satisfy the high precision requirement in the portrait segmentation problem. 

In portrait segmentation, precise segmentation around object boundaries
is crucial but challenging. For applications like background replacement, accurate and smooth boundary segmentation (such as hair and clothes) is the key for better visual effects. However, this has long been one of the most challenging part of portrait segmentation, especially when using convolutional neural networks (CNN). Since the neighborhood of boundary pixels contains a mixture of both foreground and background labels, convolutional filters fuse information of different classes, which may confuse the network when segmenting boundary pixels.
Previous CNN based semantic segmentation methods, which use either the conventional hard-label method or ignore the boundary pixels during training ~\cite{ChenPSA17} ~\cite{BadrinarayananK15}~~\cite{xz_fdnn}, fail to solve this problem. 
These methods aim to train a better model to separate foreground and background while sacrificing the accuracy when predicting the boundary pixels.



\begin{figure*}
\centering
  \includegraphics[width=.99\linewidth]{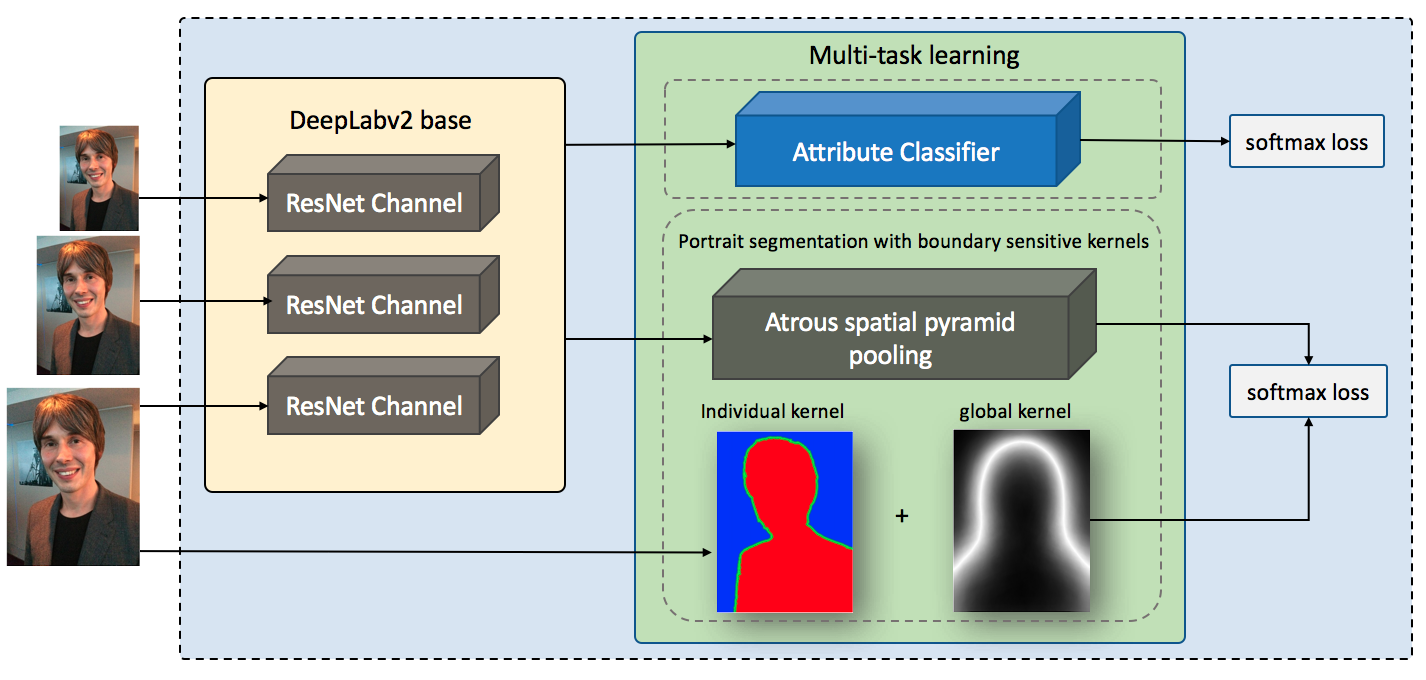}
\caption{The whole architecture of our framework.}
\label{fig:main}
\end{figure*}

In this paper, we propose a new boundary-sensitive network (BSN) for more accurate portrait segmentation. In contrast to conventional semantic image segmentation systems, we dilate the contour line of the portrait foreground and label the boundary pixels as the third class with the proposed soft-label method. Two boundary-sensitive kernels are introduced into the loss function to help the network learn better representations for the boundary class as well as govern an overall shape of the portrait. The first boundary-sensitive kernel is designed for each training image such that a floating point vector is assigned as a soft label for each pixel in the boundary class. The second boundary-sensitive kernel is a global kernel where each location in the kernel indicates the probability of the current location belonging to the boundary class. Furthermore, a boundary-sensitive attribute classifier is trained jointly with the segmentation network to reinforce the training process. We evaluate our method on PFCN \cite{pfcn}, the largest available portrait segmentation dataset. Our method achieves the best quantitative performance in mean IoU at 96.7\%. 
In order to show the effectiveness and generalization capability of our method, we further test on the portrait images collected from COCO \cite{DBLP:journals/corr/LinMBHPRDZ14}, COCO-Stuff \cite{DBLP:journals/corr/CaesarUF16a}, PASCAL VOC \cite{Everingham10} and the experiment results demonstrate that our method significantly outperforms all other state-of-the-art methods.

The rest of this paper is organized as follows: Section 2 reviews the previous work on related problems. Section 3 describes the general framework and the three boundary-sensitive techniques in detail. Section 4 discusses and analyzes the experimental results. Section 5 draws conclusions and discusses further work.

\begin{figure*}
\centering
  \includegraphics[width=.99\linewidth]{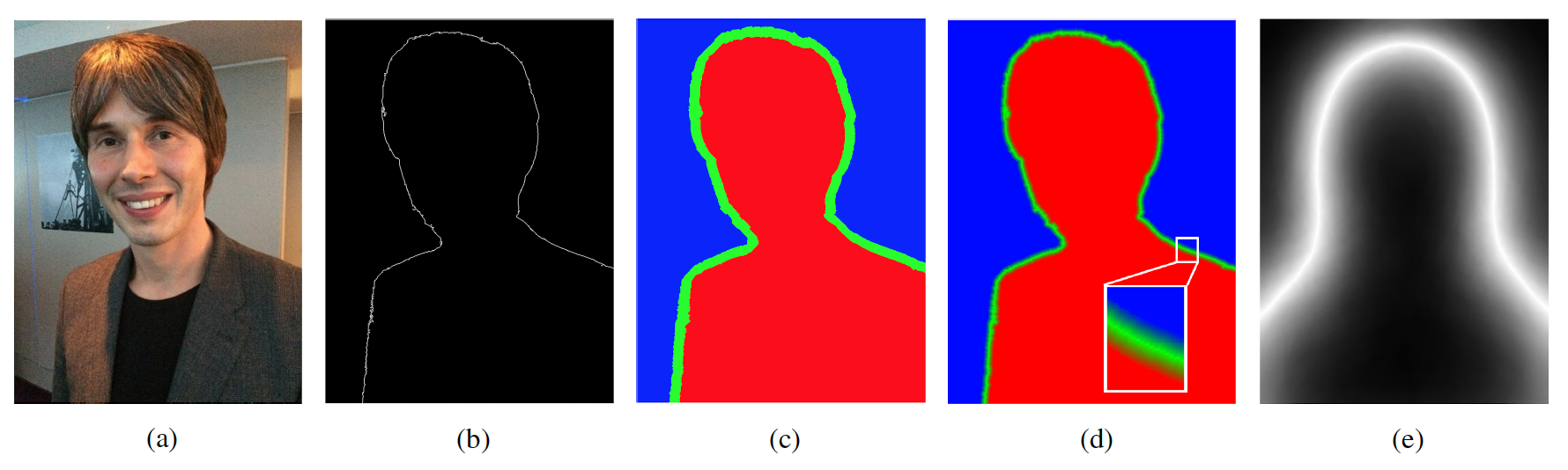}
\caption{The kernel generating process in our method: (a) represents the original image; (b) represents the detected contour line; (c) shows the three class labels: foreground, background, and dilated boundary; (d) shows the individual boundary-sensitive kernel; (e) shows the global boundary-sensitive kernel.}
\label{fig:bsmasks}
\end{figure*}

\section{Related work}
\textbf{Semantic segmentation} systems can be categorized as unsupervised methods and supervised methods. Unsupervised methods solve the semantic segmentation problem with classic machine learning techniques include thresholding, hashing, K-means clustering, topic model, graph-cut \cite{ncut}~\cite{xz_icdar}~\cite{xz_cvpr}, etc. On the other hand, conventional supervised methods treat the semantic segmentation problem as a pixel-wise classification problem which first build hand-crafted features and then train classifiers such as Support Vector Machines \cite{svm}, Random Forest \cite{Breiman2001}, etc.

In recent years, convolutional neural network (CNN) based methods have been successfully applied to semantic segmentation. In 2014, Long~\etal \cite{fcn} introduced the end-to-end Fully Convolutional Networks (FCN) which takes a natural image as input and performs dense pixel-wise predictions to generate a segmentation map of the same size as the input image. Fully connected layers are removed from this network to preserve the spatial information and deconvolutional layers are proposed for up-sampling to recover the full image size. This paradigm popularized the CNN based method and was quickly adopted by subsequent approaches. In traditional CNN architectures, pooling layer was introduced to increase the receptive field as the network goes deeper. However, it also decreases the resolution of feature map. Yu~\etal. \cite{dilated} proposed the dilated convolutional layer to replace the pooling layer, which allows for increasing the size of the receptive field without losing resolution in feature maps. Chen~\etal. \cite{CP2016Deeplab} proposed the DeepLab system which passes multiple rescaled input images to different network branches in parallel and combines the features maps with max operation at the end. 

\textbf{Portrait segmentation} is generally regarded as a sub-problem of semantic segmentation, and it is different from traditional segmentation in two aspects. First, the foreground object is limited to only people which provides additional prior information. Meanwhile, portrait segmentation has higher precision requirement on boundary area. Shen~\etal ~\cite{pfcn} fine-tuned a portrait segmentation system from a pre-trained FCN network with portrait images. To provide more portrait-specific information to the network, two normalized $x$ and $y$ position channels and one mean mask shape channel are added to the input image. Shen~\etal~\cite{shen2017high} proposed a joint correspondence and segmentation estimation method by using extra information provided by dual-lens camera. 

While most methods can easily generate a rough segmentation, they generally fail to provide precise segmentation near the object boundaries. For refining the predictions near the boundaries, the most commonly used solution is employing Conditional Random Fields (CRF) along with CNN. Deeplab\cite{CP2016Deeplab} employs dense CRF after CNN as a post processing method to smooth out the predictions. However, CRF is generally used as a post-precessing step and may be quite time-consuming.

\section{Boundary-sensitive portrait segmentation}
The architecture of our framework is shown in Figure ~\ref{fig:main}. We use DeepLabv2\_ResNet101 model as the base segmentation network. DeepLabv2\_ResNet101 consists of three ResNet101 branches at the base which process different scales of the input image. Then the three branches are followed by the atrous spatial pyramid pooling (ASPP) at different dilation rates and fused together at the end. For more details please refer to \cite{CP2016Deeplab}. To make the model more sensitive to a portrait's boundary, during training, we label the training samples with three non-overlapping classes: foreground, boundary, and background, using the soft-label method described below. One individual boundary-sensitive kernel and one global boundary-sensitive kernel are introduced when updating the loss function, which affect both the forward pass and the back-propagation. The generation process of the two kernels are shown in Figure ~\ref{fig:bsmasks}. Furthermore, an attribute classifier which shares the base layers with BSN is trained jointly with the segmentation task to reinforce the training process.

\subsection{The individual boundary-sensitive kernel and the soft-label method}
To better address the boundary prediction problem, we introduce the individual boundary-sensitive kernel. We label the boundary class as a third class to separate from foreground and background classes and assign soft-labels to pixels in the boundary class as follows. First, the portrait's contour line is identified in the ground truth segmentation map with the Canny edge detector \cite{canny}. The contour is then dilated to be P-pixels in width and that map is overlayed onto the ground truth segmentation map. We call the new label map the individual boundary-sensitive kernel. For each pixel in the kernel, a $1\times 3$ floating-point vector $K^{indv}=[l^{fg}, l^{bdry}, l^{bg}]$ is assigned as the soft-label to represent how likely the current pixel belongs to each class. The $K^{indv}$ is computed as Equations \eqref{eq:1_1} \eqref{eq:1_2} \eqref{eq:1_3}.

\begin{equation}
l^{bdry}_i=
\begin{cases}
	\cfrac{\min\limits_{\forall C_j \in C} ||I_i-C_j||}{\sum\limits_k \min\limits_{\forall C_j \in C} ||I_k-C_j||}  &\text{, if $i\in$ boundary} \\
	0 &\text{, if $i\in$ foreground}\\
    0 &\text{, if $i\in$ background}
\end{cases} 
\label{eq:1_1}
\end{equation}

\begin{equation}
l^{fg}_i=
\begin{cases}
	\mathbbm{1}(M_i\in fg)(1-l^{bdry}_i) &\text{, if $i\in$ boundary} \\
	1 &\text{, if $i\in$ foreground} \\
	0 &\text{, if $i\in$ background}
\end{cases} 
\label{eq:1_2}
\end{equation}

\begin{equation}
l^{bg}_i=
\begin{cases}
	\mathbbm{1}(M_i\in bg)(1-l^{bdry}_i) &\text{, if $i\in$ boundary} \\
	0 &\text{, if $i\in$ foreground} \\
	1 &\text{, if $i\in$ background}
\end{cases}
\label{eq:1_3}
\end{equation}

where $\min\limits_{\forall C_j \in C} ||I_i-C_j||$ represents the distance from the current pixel $I_i$ to the nearest point on the contour line $C$. $M_i$ represents the binary label of the current pixel in the original label map $M$. We can see that pixels in the foreground/background class are labeled as $[1,0,0]$/$[0,0,1]$ and pixels in the boundary class are labeled with a floating-point vector. The soft-label method computes $l^{bdry}$ as the normalized distance from the current pixel to the nearest point on the contour and sets $l^{fg}$ and $l^{bg}$ to either $(1-l^{bdry})$ or $0$ based on the class label of the current pixel in the ground truth segmentation map. During the forward pass for each pixel in one sample, the new formula for updating the loss function can be expressed as Equation \eqref{eq:1_4}:
\begin{equation} 
\epsilon=-\sum^{c}_{j=1}K^{indv}_j\times log(\frac{e^{z_j}}{\sum\limits_k e^{z_k}})=-\sum^{c}_{j=1}K^{indv}_j\times log(y_j)
\label{eq:1_4}
\end{equation}
where $l_j$ denotes the soft-label for class $j$ and $y_j=e^{z_j}/\sum\limits_k e^{z_k}$ denotes the softmax probability for this class. $c$ represents all the three classes. The new back-propagation for this sample can be derived as in Equation \eqref{eq:1_5}:

\begin{equation}
\begin{aligned}
\frac{\partial \epsilon}{\partial z_i}&=-\sum^{c}_{j=1}K^{indv}_j\times \frac{\partial log(y_j)}{\partial z_i} \\
&=-(\frac{K^{indv}_i}{y_i}\times \frac{\partial y_i}{\partial z_i}+\sum^c_{j\neq i}\times \frac{K^{indv}_j}{y_j}\frac{\partial y_j}{\partial z_i})\\
&=-(\frac{K^{indv}_i}{y_i}\times y_i\times (1-y_i)-\sum^c_{j\neq i}\frac{K^{indv}_j}{y_j}\times (y_j\times y_i))\\
&=-(K^{indv}_i-y_i\times \sum^c_{j=1}l_j)=-(K^{indv}_i-y_i)
\end{aligned}
\label{eq:1_5}
\end{equation}
The last step holds since the soft-label vector sums to one.

By using the soft-label method, we can see that boundary pixels contribute not only to the boundary class but also to the foreground/background class in a weighted manner based on how close it is to the contour line. 


\subsection{The global boundary-sensitive kernel}
By the nature of aligned portrait images, it is likely that some locations in the image, such as the upper corner pixels, should belong to the background with very high probabilities while some other locations, such as the middle bottom pixels, should belong to the foreground with high probabilities. These pixels should be more easily classified, while pixels in between should be harder to classify. We estimate a position sensitive prior from the training data.


We design a global boundary-sensitive kernel to guide the network to learn a better shape prediction specifically for portrait images. The global kernel is designed as follows. First, a mean mask $\overline{M}$ is computed using the average of all ground truth segmentation maps from the training samples. This generates a probability map where the value at each location indicates how likely the current location belongs to foreground/background. Second, Equation \eqref{eq:2_1} is employed to generate the global boundary-sensitive kernel. All the values are normalized to range $[a, b]$. A larger value close to $b$ in the global kernel indicates that the current location has a higher probability to be boundary. In other words, this location should be more difficult for the network to classify. To force the network to focus more on the possible boundary locations, we weight the locations with their corresponding kernel values when updating the loss function. When performing the forward pass for one pixel location in one sample, we update the loss function as equations \eqref{eq:2_2}

\begin{equation} \label{eq:2_1}
K^{global}=b-(1-\frac{|\overline{M}-0.5|}{0.5})\times (b-a)
\end{equation}

\begin{equation} \label{eq:2_2}
\epsilon\mathrel{{-}{=}}K^{global}_{s}\times \sum_{j}\mathbbm{1}(j=c)\times log(y_j)
\end{equation}
where $K^{global}_{s}$ denotes the global kernel value at the pixel location $s$. $g$ denotes the ground truth class label for the current pixel location. During back-propagation, the new gradient is computed as Equation \eqref{eq:2_3}:
\begin{equation} \label{eq:2_3}
\begin{aligned}
\frac{\partial \epsilon}{\partial z_i}&=-K^{global}_{s}\times \sum\limits_j\mathbbm{1}(j=g)\times \frac{\partial log(y_j)}{\partial z_i} \\
&=-K^{global}_{s}\times(\frac{1}{y_i}\times \frac{\partial y_i}{\partial z_i})\\
&=-K^{global}_{s}\times(\frac{1}{y_i}\times y_i\times (\mathbbm{1}(i=g)-y_i))\\
&=-K^{global}_{s}\times(\mathbbm{1}(i=g)-y_i)
\end{aligned}
\end{equation}

From the new forward pass and back-propagation functions we can see that the pixels that are more likely to be located in the boundary (\eg, the pixels lying within the brighter region in Figure ~\ref{fig:bsmasks} (e)) are weighted higher so that they contribute more to the loss. This guides the network to be more sensitive to the difficult locations.

\subsection{The boundary-sensitive attribute classifier}
\begin{figure}[htbp]
\center
\begin{minipage}[t]{0.45\linewidth}
    \includegraphics[width=0.95\linewidth]{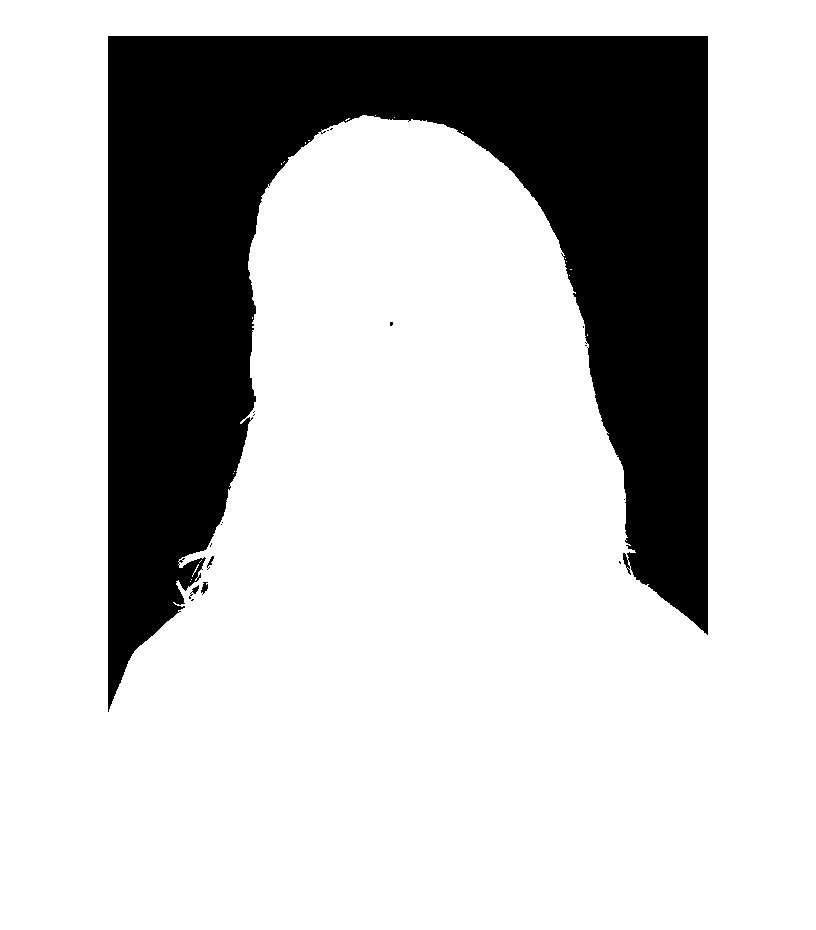}
\end{minipage}%
\begin{minipage}[t]{0.45\linewidth}
    \includegraphics[width=0.95\linewidth]{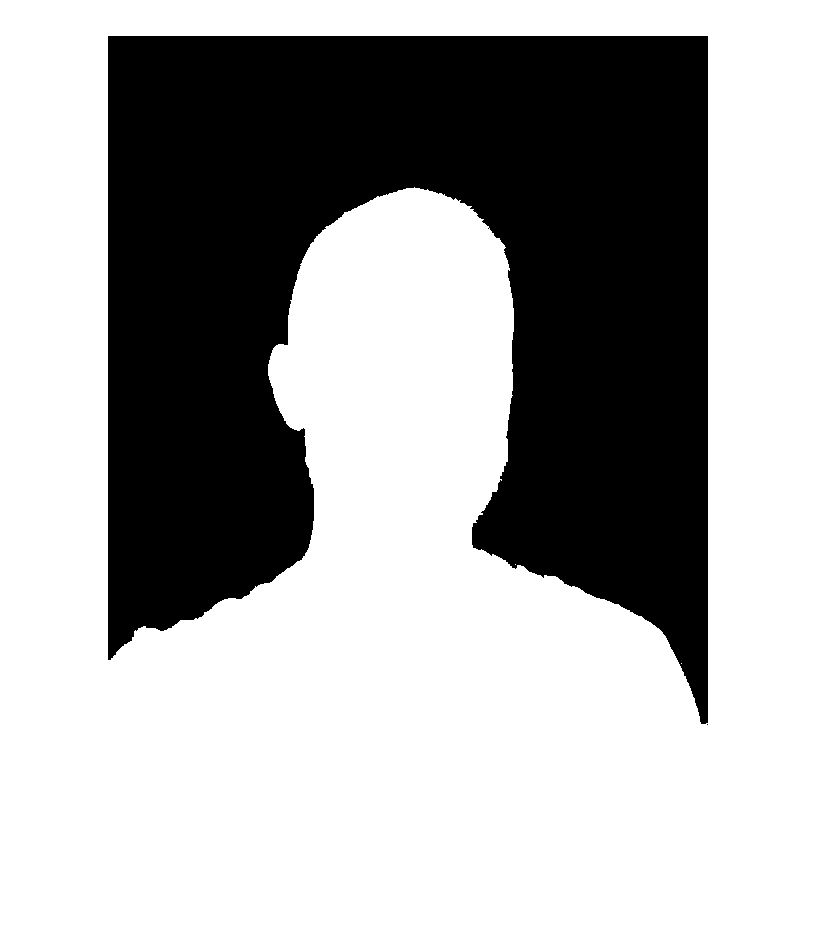}
\end{minipage}
\caption{An example of how boundary-sensitive attributes affect the portrait's shape: long hair vs. short hair.}
\label{fig:img_attr}
\end{figure}

Portrait attributes such as long/short hair play an important role in determining a portrait's shape. Training a network which is capable of classifying boundary-sensitive attributes will give more prior information to the system, which further makes the system more accurate and efficient on boundary prediction. Motivated by this idea, we train an attribute classifier jointly with the portrait segmentation network for multi-task learning. An example of how the hair style attribute changes the boundary shape is shown in Figure ~\ref{fig:img_attr}.

To design the attribute classifier, the base layers from ``conv1\_1" to ``pool5" are shared between the segmentation network and the classifier. Above this, for each channel, we add three more fully connected layers. The first two fully connected layers have $1024$ neurons and are followed by a dropout layer and a ReLU layer. The last fully connected layer has two neurons for binary classification.

\section{Experiments and results analysis}
\subsection{Training settings and evaluation settings}
\textbf{Model details:} To train our portrait segmentation system, we fine-tune the DeepLabv2\_ResNet101 model using the training set of the PFCN dataset. We will introduce this dataset in the next subsection. There are three ResNet branches in DeepLabv2. In each branch, $4$ atrous convolution layers are added in parallel with dilation factors $[6,12,18,24]$ and then summed together to produce the final feature map. Element-wise max operation is performed at the end over the three branches to produce the final prediction. To generate the individual kernel, we dilate the contour line to 10-pixels in width and label the dilated boundary using the soft-label method. We select the weight range in the global kernel as $[0.9,1]$. Following PortraitFCN+, in addition to the three RGB channels, we add two normalized $x$ and $y$ position channels and one mean mask shape channel into the input. For more details please refer to  \cite{pfcn}. At each iteration, a random patch of size $400\times 400$ is cropped out from the original image and randomly flipped with probability $0.5$ for data augmentation. Then the input image is rescaled by factors of $[0.5,0.75,1.0]$ as the new input images to the three branches of the DeepLabv2 network. To train the attribute classifier, we label the training images into long/short hair classes. We use Stochastic Gradient Descent (SGD) with a learning rate of $2.5e^{-4}$ to train the model for 20K iterations without the attribute classifier. Then we decrease the learning rate by a factor of $10$ and add the attribute classifier to train the model for another 20K iterations. The whole network is built with the Caffe deep learning framework \cite{jia2014caffe}.

During testing, we ignore the boundary class and the attribute classifier. Only probabilities from foreground and background classes are used for segmentation. 

\textbf{Mean IoU:} The standard mean Intersection-over-Union (IoU) metric is used to evaluate the segmentation performance. The mean IoU is computed as following.

\begin{equation}
\overline{\text{IoU}} = \frac{1}{N}\times \sum\limits_i^N \frac{A^{seg}_i\cap A^{gt}_i}{A^{seg}_i\cup A^{gt}_i}
\label{eq:3}
\end{equation}
where $A^{seg}_i$ and $A^{gt}_i$ represent the area of the segmentation results and the ground-truth label mask for the $i_{th}$ testing sample, respectively.


\subsection{Results on the PFCN dataset}
\begin{figure}[htbp]
\centering
    \includegraphics[width=0.99\linewidth]{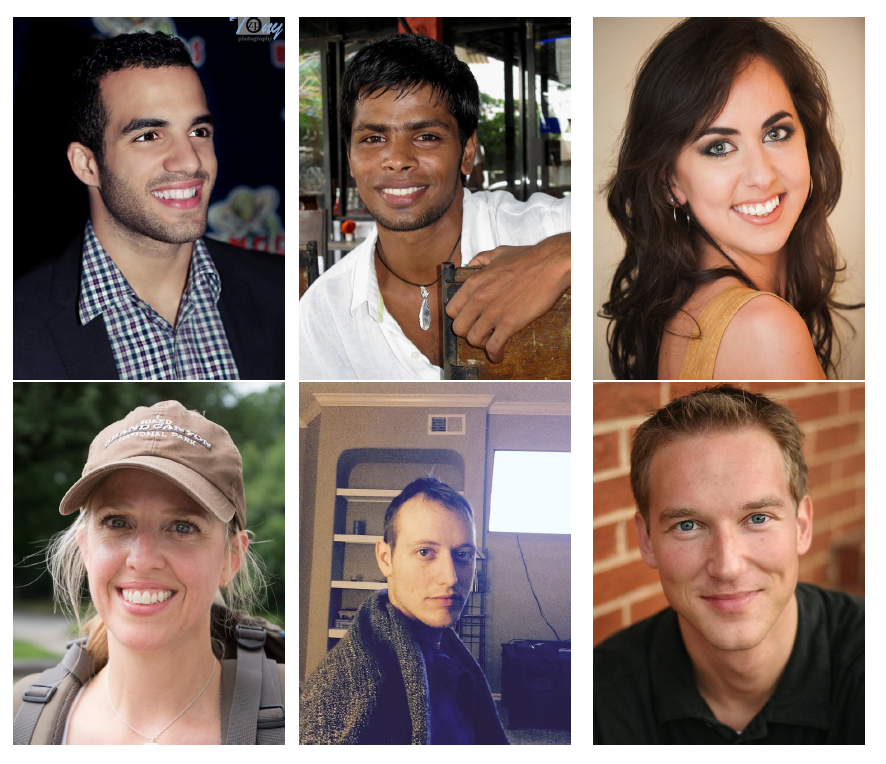}
\caption{Sample images from the PFCN dataset.}
\label{fig:pfcn}
\end{figure}

We evaluate the proposed method on the largest publicly available portrait segmentation dataset \cite{pfcn}. This dataset is collected from Flickr and manually labeled with variations in age, pose, appearance, background, lighting condition, hair style, accessory, etc. Most of the portrait images are captured by the frontal cameras of mobile phones. This dataset consists of $1800$ portrait images which are split into $1500$ training images and $300$ testing images. All the images are scaled and cropped into size $800\times 600$. In one portrait image, the pixels are labeled as either ``foreground" or ``background". We will refer to this dataset as PFCN dataset. Some sample images from the PFCN dataset are given in Figure ~\ref{fig:pfcn}. 

\begin{figure}[htbp]
\centering
    \includegraphics[width=0.99\linewidth]{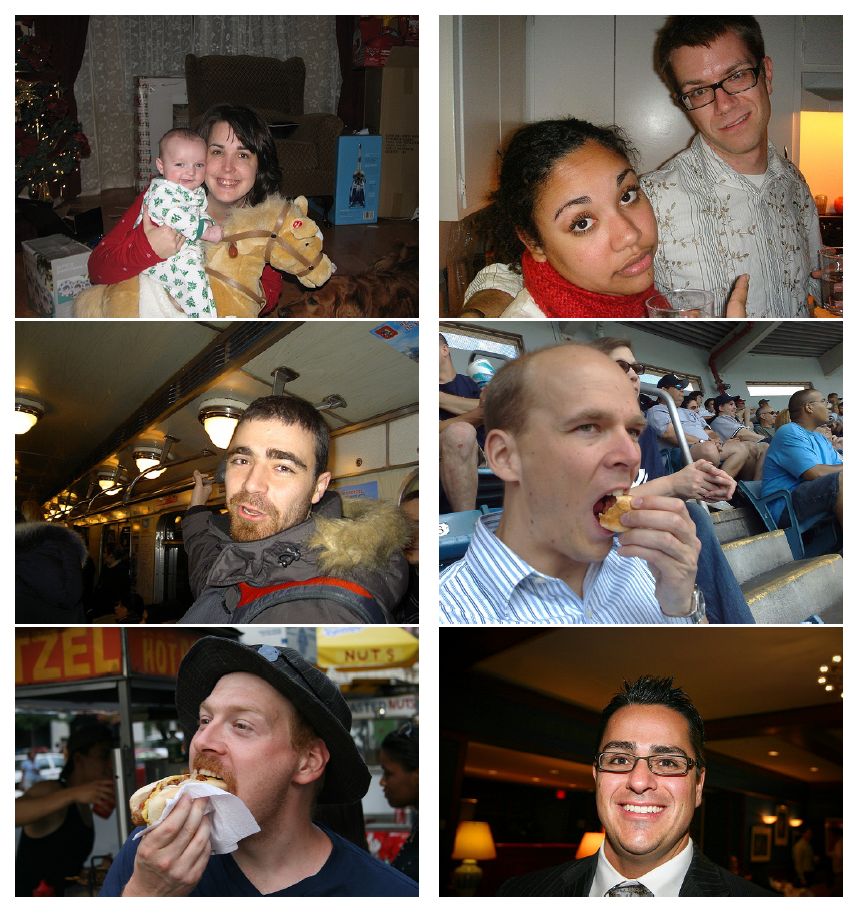}
\caption{Sample images from COCO, COCO-Stuff, and Pascal VOC portrait datasets.}
\label{fig:pascal}
\end{figure}

We compare with the state-of-the-art method reported on this dataset: PortraitFCN+~\cite{pfcn} and the DeepLabv2\_ResNet101 fine-tuned model , which we will refer to as PortraitDeepLabv2. The PortraitDeepLabv2 model is fine-tuned using the same $6$-channel training data as PortraitFCN+ and the same training settings as BSN. For ablation study, we report results of four models from our work: train with the attribute classifier only (BSN\_AC), train with the global boundary-sensitive kernel only (BSN\_GK), train with the individual boundary-sensitive kernel only (BSN\_IK), and the all-in-one model (BSN). Our final model achieves the state-of-the-art mean IoU at 96.7\%. The quantitative result comparison is given in Table ~\ref{tab:1}. Result from graph-cut~\cite{ncut} is shown as the baseline.

\begin{table}
\begin{center}
\begin{tabular}{|c|c|}
\hline
Method & Mean IoU \\
\hline\hline
Graph-cut & 80.0\% \\
PortraitFCN+ & 95.9\% \\
PortraitDeepLabv2 & 96.1\% \\
BSN\_AC (ours) & 96.2\% \\
BSN\_GK (ours)& 96.2\% \\
BSN\_IK (ours)& 96.5\% \\ 
BSN (ours)& \textbf{96.7\%} \\
\hline
\end{tabular}
\end{center}
\caption{Quantitative performance comparisons on the PFCN dataset.}
\label{tab:1}
\end{table}

\begin{figure*}
\centering
    \includegraphics[width=0.99\linewidth]{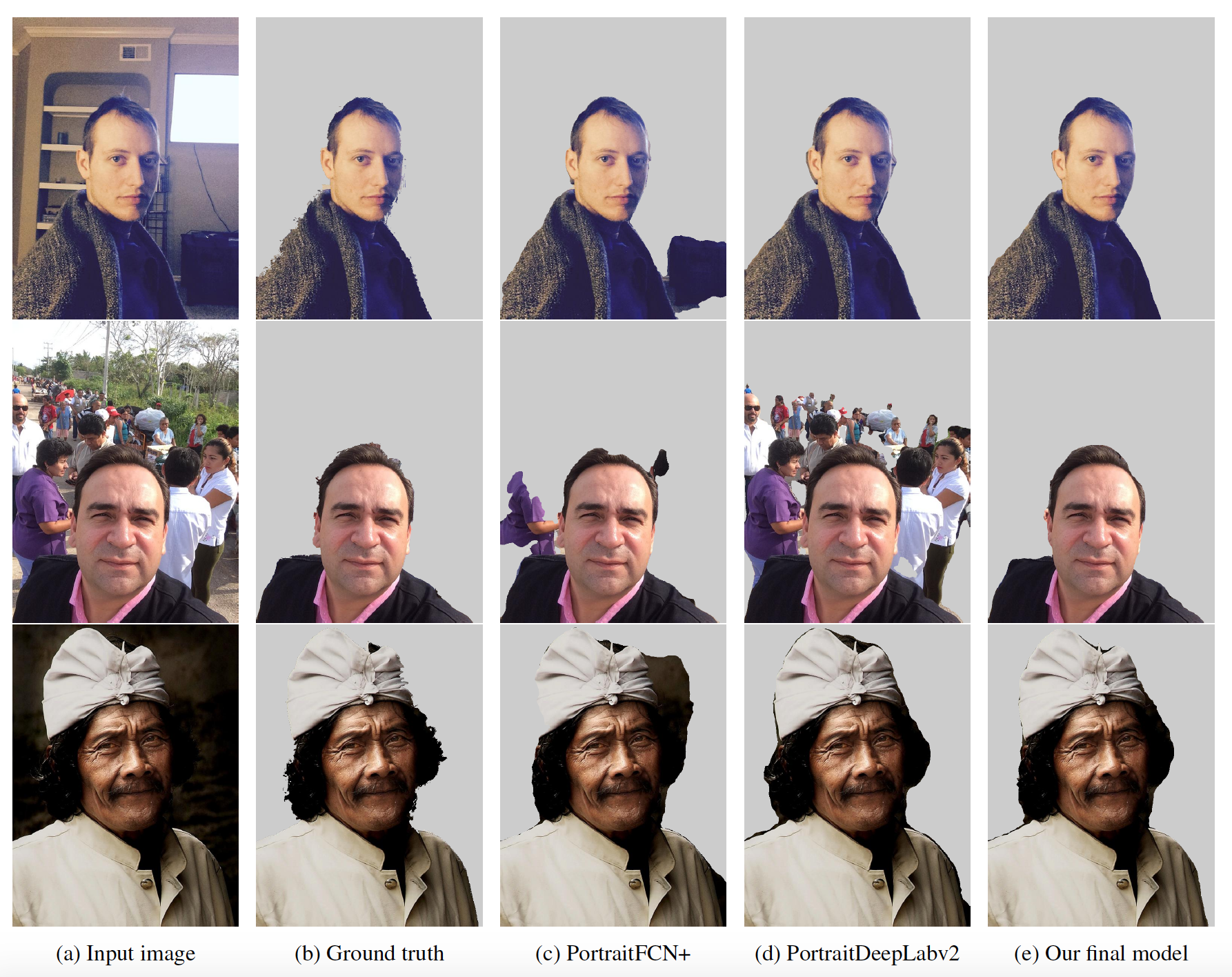}
\caption{Result visualizations of three challenging examples. The first row shows contains confusing objects in the background; the second row includes multiple people in the background; in the third row the background color is close to the foreground.}
\label{fig:challenge}
\end{figure*}

\subsection{Evaluation on other datasets: }
Since the performance on the PFCN dataset is pretty high and data for the boundary class is unbalanced compare to foreground/background, a good performance on boundary segmentation may only lead to marginal improvement in mean IoU on this dataset. Thus we further test our method on the portrait images collected from three more popular semantic segmentation datasets to evaluate the effectiveness of our boundary-sensitive techniques.

\textbf{COCO portrait: }We automatically collect all the portrait and portrait-like images from the COCO dataset. We run a face detector over the dataset and keep the images only containing one person where the face area covers at least 10\% of the whole image. There are 626 images in total with ground truth segmentation maps. We will refer to this dataset as COCO portrait. COCO portrait is more challenging than the PFCN data in various ways such as large pose variations, large occlusions, unlabeled individuals appear on the background, large portion of background, different kinds of accessories, etc. 

\textbf{COCO-Stuff portrait: }The COCO-Stuff dataset augments the COCO dataset with refined pixel-level stuff annotations on $10K$ images. We collect 92 portrait and portrait-like images from this dataset. The quality of images in this dataset are same as COCO portrait. We will refer to this dataset as COCO-Stuff portrait.

\textbf{Pascal VOC portrait: }We use the same method to collect portrait and portrait-like images from the Pascal VOC 2007, 2008, and 2012 datasets. Due to the lack of ground truth segmentation maps on this dataset, 62 images are collected. The images in this dataset are also challenging and unconstrained. We will refer to this dataset as PASCAL VOC portrait. Some sample images from the three datasets are illustrated in Figure ~\ref{fig:pascal} and the statistics are given in Table ~\ref{tab:2_1}.

To test the generalization capability of our model, we directly test on these three datasets without fine-tuning. We achieve 77.7\% mean IoU, 72.0\% mean IoU, and 75.6\% mean IoU on COCO portrait, COCO-stuff portrait, and PASCAL VOC portrait, respectively. We significantly outperform PortraitFCN+ on all the three datasets. The result comparisons are illustrated in Table ~\ref{tab:2}. Since the DeepLabv2 model is trained on these dataset, we can not compare with it directly.

\begin{table}
\begin{center}
\begin{tabular}{|c|c|}
\hline
Dataset  & Num. of Portrait\\
\hline\hline
COCO portrait & 626 \\
COCO-Stuff portrait & 92\\
PASCAL VOC portrait & 62\\
\hline
\end{tabular}
\end{center}
\caption{Statistics of the three portrait datasets.}
\label{tab:2_1}
\end{table}

\begin{table}
\begin{center}
\begin{tabular}{|c|c|c|c|}
\hline
Method  & COCO & COCO- & PASCAL \\ & & Stuff & VOC\\
\hline\hline
PortraitFCN+  & 68.6\% & 60.8\% & 59.5\%\\
BSN (ours) & \textbf{77.7}\% &\textbf{72.0\%} & \textbf{75.6\%}\\
\hline
\end{tabular}
\end{center}
\caption{Quantitative performance comparisons on COCO portrait, COCO-Stuff portrait and Pascal VOC portrait datasets.}
\label{tab:2}
\end{table}

\subsection{Result analysis}
\subsubsection{Results visualization on challenging scenarios}
We visualize the overall performance of our BSN model compared to DeepLabv2 and PortraitFCN+ using three challenging scenarios: confusing objects in the background, multiple people appear in the image, and the background color theme is close to the foreground. Figure ~\ref{fig:challenge} shows that our model is more accurate and robust than other methods even under challenging conditions.

\subsubsection{Accurate boundary segmentation}
\begin{figure}[htbp]
\centering
    \includegraphics[width=0.99\linewidth]{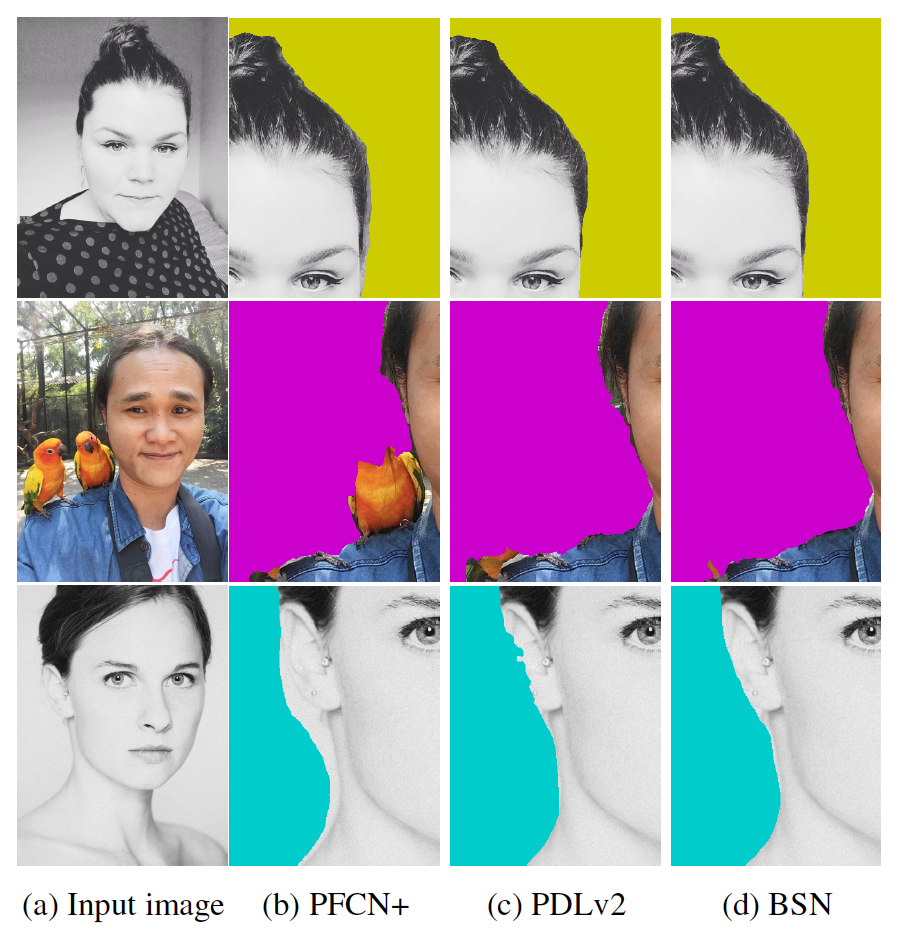}
\caption{Boundary segmentation comparisons. The first column are the original images.  The three subsequent columns represent the results from the PortraitFCN+ method, the fine-tuned DeepLabv2 model with the attribute classifier, and our final model (magnified for best viewing).}
\label{fig:img_bdryseg}
\end{figure}

Our method also delivers more precise boundary predictions thanks to its novel boundary-sensitive segmentation techniques. Figure ~\ref{fig:img_bdryseg} shows the comparison of our method with DeepLabv2 and PortraitFCN+ in three challenging scenarios: hair segmentation, accessory segmentation and ear segmentation. Results reveal that while other methods have difficulty in segmenting accessories and small body parts, our method can provide a smooth and accurate segmentation.

\begin{figure}[htbp]
\centering
    \includegraphics[width=0.99\linewidth]{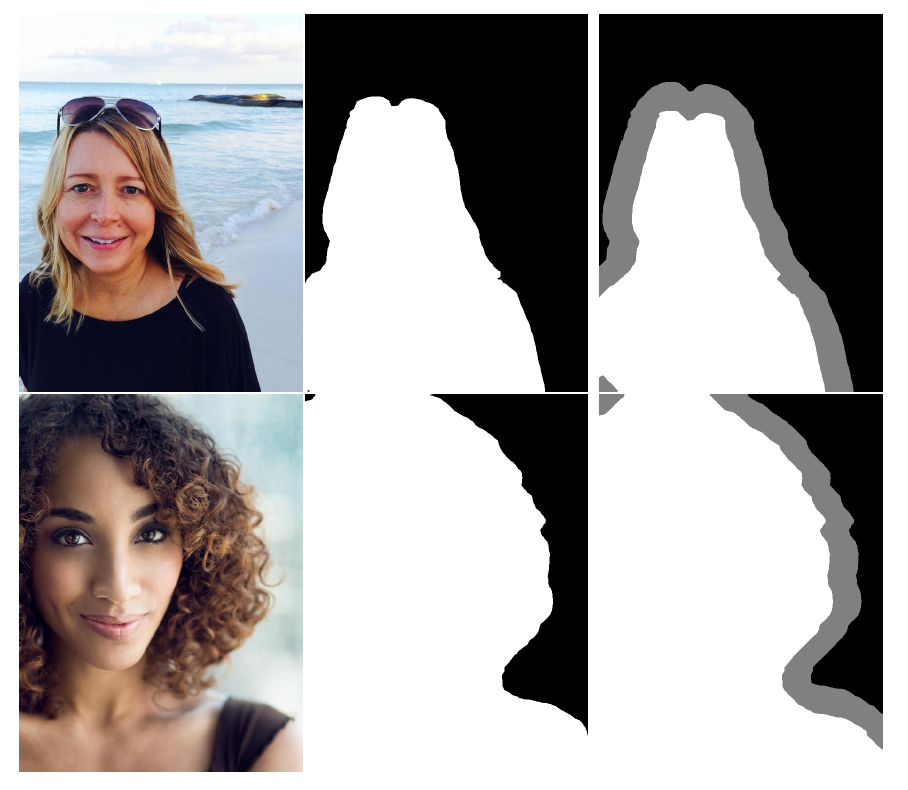}
\caption{Trimaps generated from our segmentation maps.}
\label{fig:img_trimap}
\end{figure}

\subsubsection{Generating trimap for image matting}
Since our method can deliver an accurate boundary prediction, it is a natural extension to generate trimaps for image matting models. After performing segmentation, we use the same technique during training to dilate the boundary pixels to 10-pixels in width. Several examples are shown in Figure ~\ref{fig:img_trimap}.

\subsubsection{Applications of portrait segmentation}
\begin{figure}[htbp]
\centering
    \includegraphics[width=0.99\linewidth]{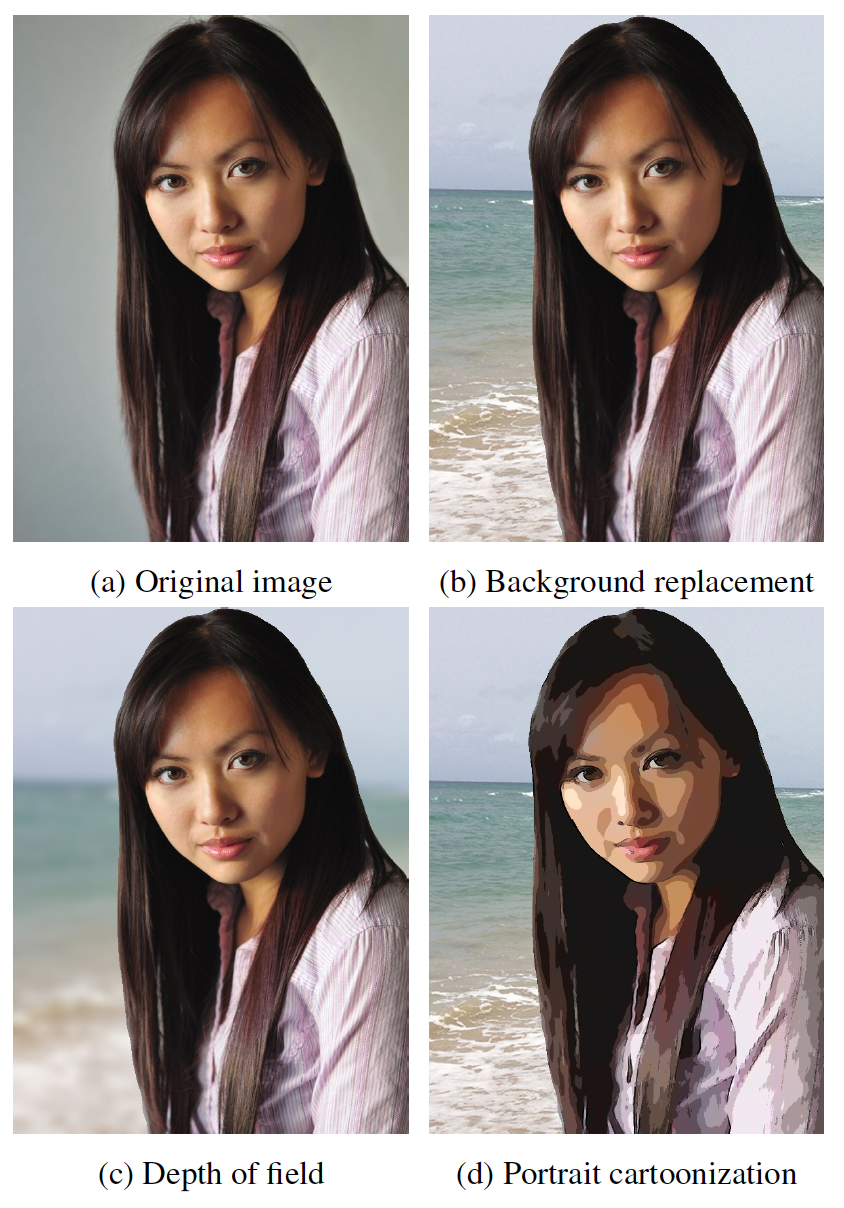}
\caption{Some applications of portrait segmentation.}
\label{fig:img_app}
\end{figure}

Portrait segmentation has been widely used in various image processing applications such as background replacement, depth of field, augmented reality, image cartoonization, etc. We show some applications in Figure ~\ref{fig:img_app}. 

\section{Conclusion and discussion}
We present a boundary-sensitive portrait segmentation system. Two boundary-sensitive kernels are introduced into the loss function. One gives more boundary information for each individual image and one governs the overall shape of portrait prediction. An attribute classifier is trained jointly with the segmentation network to reinforce the training process. Experiments are conducted on the largest publicly available portrait segmentation dataset as well as portrait images collected from other three popular semantic segmentation datasets. We outperform the previous state-of-the-arts in both quantitative performance and visual performance.

For future work, we would like to extend our boundary-sensitive methods to general semantic segmentation problem and explore more semantic attributes to reinforce the training process.

{\small
\bibliographystyle{ieee}
\bibliography{egbib}
}
\end{document}